\documentclass{article}

\usepackage{PRIMEarxiv}

\usepackage{graphicx}
\usepackage{booktabs}
\usepackage{multirow}
\usepackage{multicol}
\usepackage{makecell}
\usepackage{epsfig}
\usepackage[utf8]{inputenc} 
\usepackage[T1]{fontenc}    
\usepackage{amsmath}
\usepackage{adjustbox}
\usepackage{hyperref}       
\usepackage[capitalize]{cleveref}
\usepackage{url}            
\usepackage{booktabs}       
\usepackage{amsfonts}       
\usepackage{nicefrac}       
\usepackage{microtype}      
\usepackage{lipsum}
\usepackage{fancyhdr}       
\usepackage{graphicx}       
\graphicspath{{media/}}     
\usepackage{enumitem}

\usepackage{algorithm}
\usepackage{algorithmicx}
\usepackage{algpseudocode}
\usepackage{bm}

\usepackage{rotating}
\usepackage{gensymb}
\usepackage{wrapfig}
\usepackage{xcolor}
\usepackage{array}
\usepackage{colortbl}
\usepackage{pifont}
\definecolor{mygray}{gray}{.92}

\DeclareMathDelimiter{(}{\mathopen} {operators}{"28}{largesymbols}{"00}
\DeclareMathDelimiter{)}{\mathclose}{operators}{"29}{largesymbols}{"01}

\newcommand{\model}{EvSNet\xspace}
\newcommand{\method}{EventTracer\xspace}
\newcommand{\dataset}{ETScenes\xspace}

\definecolor{blueevent}{RGB}{58, 89, 209}
\definecolor{redevent}{RGB}{209, 89, 58}
\definecolor{isb}{RGB}{124, 134, 65}
\definecolor{sb}{RGB}{128, 53, 14}

\usepackage{xspace}

\title{EventTracer: Fast Path Tracing-based Event Stream Rendering}

\author{
  Zhenyang Li\textsuperscript{1}\thanks{These authors contributed equally.},
  Xiaoyang Bai\textsuperscript{1}\footnotemark[1],
  Jinfan Lu\textsuperscript{1,2},
  Pengfei Shen\textsuperscript{1},
  Edmund Y. Lam\textsuperscript{1},
  Yifan Peng\textsuperscript{1}\thanks{e-mail: evanpeng\texttt{@}hku.hk} \\[1ex]
  \scriptsize
  \textsuperscript{1}The University of Hong Kong\quad
  \textsuperscript{2}Tsinghua University
}

\begin{document}
\maketitle

\begin{abstract}
    Simulating event streams from 3D scenes has become a common practice in event-based vision research, as it meets the demand for large-scale, high temporal frequency data without setting up expensive hardware devices or undertaking extensive data collections. 
    Yet existing methods in this direction typically work with noiseless RGB frames that are costly to render, and therefore they can only achieve a temporal resolution equivalent to 100$\sim$300 FPS, far lower than that of real-world event data.
    In this work, we propose \emph{\method}, a path tracing-based rendering pipeline that simulates high-fidelity event sequences from complex 3D scenes in an efficient and physics-aware manner.
    Specifically, we speed up the rendering process 
    via low sample-per-pixel (SPP) path tracing, and train a lightweight event spiking network to denoise the resulting RGB videos into realistic event sequences. 
    To capture the physical properties of event streams, the network is equipped with a bipolar leaky integrate-and-fired (BiLIF) spiking unit and trained with a bidirectional earth mover's distance (EMD) loss.
    Our \method pipeline runs at a speed of $\sim$4 minutes per second of 720p video, and it inherits the merit of accurate spatiotemporal modeling from its path tracing backbone.
    We show in two downstream tasks that \method captures better scene details and demonstrates a greater similarity to real-world event data than other event simulators, which establishes its as a promising tool for creating large-scale event-RGB datasets at a low cost, narrowing the sim-to-real gap in event-based vision, and boosting various application scenarios such as robotics, autonomous driving, and VR/AR.
\end{abstract}


\begin{figure}
    \centering
    \includegraphics[width=\linewidth]{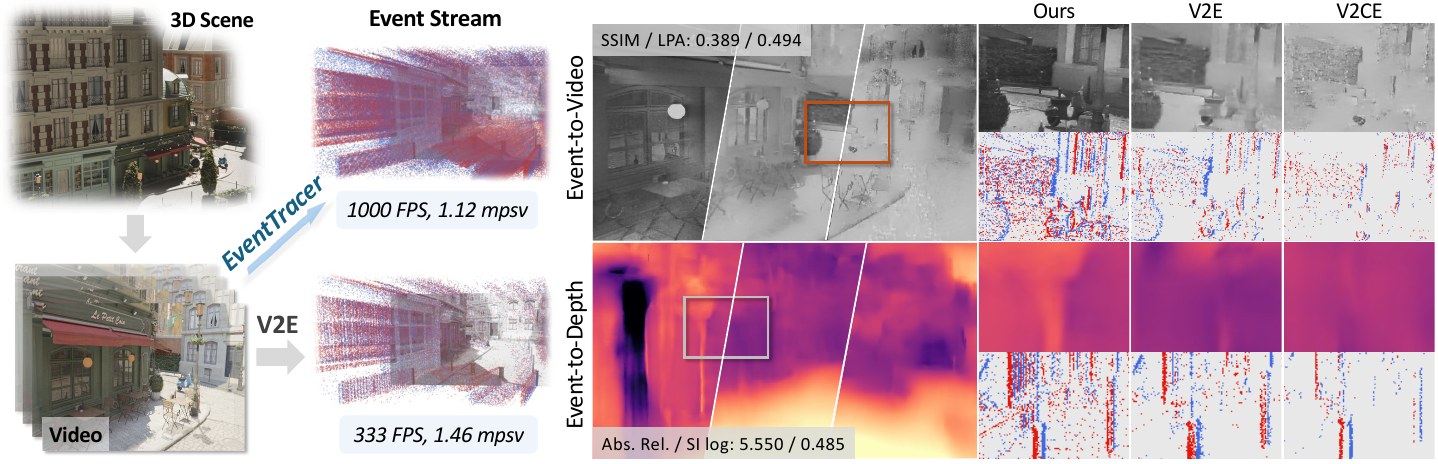}
    \caption{\textbf{Left}: Deformation-based 4DGS encounters difficulties in reconstructing scenes and rendering novel views under challenging conditions, such as significant motion and other complex dynamics. Our \model exhibits commendable performance on given scenes (e.g., ``cut-roasted-beef''). \textbf{Middle}: We compare the ground-truth optical flow, the deformation network of baseline (4DGS), and the velocity field rendered by our method. \textbf{Right}: We render the velocity field for scene Gaussians, constrain it with flow-based losses, and employ the FAD strategy to add Gaussians for dynamic objects in the canonical space.}
    \label{fig:teaser}
\end{figure}

\section{Introduction} 
\label{sec:intro}

Over the past few years, event streams have emerged as a unique modality that compensates the conventional RGB videos with high temporal resolution and motion sensitivity~\cite{gallegoEventbasedVisionSurvey2022}. In the fields of drone navigation, autonomous driving, robotics, and AR/VR, event sensory have been widely integrated into the perception and decision pipeline to enhance its performance in dynamic and challenging environments~\cite{angelopoulos2020event,xu2023taming,gehrig2024low}.
Nowadays, event-based vision is dominated by learning-based methods~\cite{zheng2023deep}, which effectively tackle the noisy nature of event streams. Yet training those models poses a high demand for large-scale event-RGB datasets, whose availability is still far from sufficient.

Compared to unimodal RGB videos, event-RGB data are noticeably more costly and time-consuming to collect. Commercial event cameras are usually expensive, and they either lack frame-based imaging capability (\textit{e.g.} Prophesee's EVK series and iniVation's DVXplorer) or feature low resolution and inferior imaging fidelity~\cite{brandli2014240}. While a few datasets have been curated using calibrated and synchronized multi-camera systems~\cite{wang2023visevent,duan2025eventaid}, they often fail to address the demand for task-specific annotations, such as depth maps, surface normals, and segmentation masks. Including those modalities will necessitate extra hardware setups or algorithmic post-processing, which further complicates the data collection pipeline.

To address this issue, prior studies~\cite{huV2eVideoFrames2021,zhangV2CEVideoContinuous2024} have explored synthetic event generation from RGB videos, requiring clear intensity frames as input. Such a constraint limits their temporal resolution to equivalently 100$\sim$300 FPS, even coupled with frame interpolation algorithms, considerably lower than the one of actual event cameras ($\geq$1,000 FPS). 
While a higher frame rate is attainable by running simulations on videos rendered from 3D scenes~\cite{rebecqESIMOpenEvent2018,  hanPhysicalBasedEventCamera2024, liBlinkFlowDatasetPush2024}, such an approach is yet limited by the tradeoff between speed and quality. On one end sits rasterization, which renders at an impeccable speed but lacks photorealism, especially under complex lighting conditions; conversely, the time consumption of noise-free global illumination techniques such as path tracing can pile up to several seconds per frame, making the entire process intolerably slow. In a word, efficient, realistic, and high temporal resolution event simulation is yet an unsolved problem.


In light of this, we propose \emph{\method}, a path tracing-based rendering pipeline that satisfies all three requirements. To speed up the rendering process, we apply Monte Carlo path tracing with a low sample-per-pixel (SPP) setting. Next, we train a \emph{lightweight event spiking network (\model)} to emulate event signals from noisy pixel illuminances. To capture the physical properties of event sensors, we propose a bipolar leaky integrate-and-fire (BiLIF) unit to activate the outputs of \model into event pulses, and train the whole network with a bidirectional earth mover's distance (EMD) loss. Unlike other learning-based approaches~\cite{zhangV2CEVideoContinuous2024}, the spatial and temporal localness of \model allows us to integrate it into Falcor~\cite{Kallweit22} using NVIDIA TensorRT~\cite{tensorrt} and fully parallelize the rendering process. The overall \method pipeline is able to simulate event streams from dynamic 3D scenes at 1,000 FPS with a speed of 1.12 minutes per second of 360p video (Fig.~\ref{fig:teaser} \textit{left}), surpassing previous works in both runtime and temporal resolution. Besides, its outputs are high dynamic range and easily extended by extra annotations such as segmentation masks and depth maps, thanks to the path tracing-based pipeline.

To evaluate the effectiveness of \method, we use it to build an event-RGB dataset, named \dataset, that consists of 90 seconds of videos rendered from 6 different 3D scenes. Using this dataset, we perform \emph{Real2Sim} 
comparison between \method, V2E~\cite{huV2eVideoFrames2021}, and V2CE~\cite{zhangV2CEVideoContinuous2024} on two downstream tasks, event-to-video reconstruction~\cite{ercan2023evreal} and event-based depth estimation~\cite{hidalgo2020learning}. Experiment results demonstrate that our simulated \dataset is more similar to real-captured data than its peers, validating the superior render quality of \method, in addition to its efficiency (Fig.~\ref{fig:teaser} \textit{right}).
Therefore, \method can serve as a powerful tool for efficiently generating high-quality synthetic event-RGB data in the near future, meeting the needs of various event-based vision tasks and narrowing the sim-to-real gap that bottlenecks the research community.
In summary, our main technical contributions include:
\begin{itemize}[leftmargin = 18pt]
    \item We propose a path tracing-based event rendering framework, dubbed \emph{\method}, that realizes fast and accurate simulation of event-RGB data from 3D scenes.
    \item We design \emph{\model}, a lightweight physics-aware neural network that can be seamlessly integrated with the path tracer.
    \item We validate \method by constructing the \emph{\dataset} dataset and testing it on different downstream vision tasks. It demonstrates improved quality and similarity to real-world data compared to those produced by existing event simulators.
\end{itemize}
\vspace{-6pt}
\section{Related Work} 
\label{sec:related}

\subsection{Event Data Collection and Synthesis}

Event cameras are novel sensory systems that asynchronously measure per-pixel illuminance variations~\cite{lichtsteiner2008128},
well-suited for challenging scenarios involving high-speed motion or power constraints.
Recently, their application has extended beyond the conventional high-speed imaging~\cite{ieng2017event, zou2021learning} into 3D vision~\cite{yu2024eventps}, intelligent camera control~\cite{louAllinFocusImagingEvent2023,lin2024embodied}, and event-based optics~\cite{shah2024codedevents}, which increasingly levitates its prospect in fields such as robotics, autonomous driving, and wearable electronics~\cite{gallegoEventbasedVisionSurvey2022}. 
With this trend, numerous datasets with paired RGB-event data have been released, including CED~\cite{scheerlinck2019ced}, LED \cite{duanLEDLargescaleRealworld2024}, and EventAid~\cite{duan2025eventaid}. 
Some datasets also include additional annotations, for example, depth maps in MVSEC \cite{zhuMultiVehicleStereo2018} and segmentation masks in EV-IMO~\cite{mitrokhin2019ev}. 
Alternative approaches like CIFAR10-DVS \cite{liCIFAR10DVSEventStreamDataset2017} image RGB videos played on high refresh rate screens with event cameras, which saves the effort of multi-camera calibration and synchronization. Yet all these datasets require costly hardware setups and time-consuming collection processes, and they are often limited in resolution, scene diversity, and video length, potentially hindering their effectiveness for downstream tasks.

Given this, there has been a growing interest in synthetic event generation, which can be categorized into two types based on: \emph{existing video sequences} and \emph{3D simulated scenes}.
In the former category, \cite{huV2eVideoFrames2021} introduced V2E, a toolbox that synthesizes realistic DVS events from gray-scale videos which models various aspects of actual event sensors. 
More recently, \cite{zhangV2CEVideoContinuous2024} developed V2CE as a learning-based tool to generate continuous event streams from videos. 
However, the temporal resolution of these video-based methods is limited by the relatively low frame rate of input videos (typically $\leq$60 FPS). Even with the help from pretrained frame interpolation models~\cite{jiang2018super}, the final temporal sensitivity still struggles to meet the needs of downstream applications.

Conversely, simulation-based methods leverage physically accurate 3D models to generate synthetic event data under fully controlled conditions. 
Notable works include ESIM~\cite{rebecqESIMOpenEvent2018}, BlinkFlow \cite{liBlinkFlowDatasetPush2024}, and PECS~\cite{hanPhysicalBasedEventCamera2024}. Although these simulators deliver high temporal resolution results that are easily extended by additional annotations, their efficiency is constrained by the backbone rendering speed. For example, noiseless (2,048 SPP) path tracing runs at 5.42 seconds per 720p frame, suggesting 1.5 hours of runtime for rendering 1,000 FPS videos and subsequently simulating high-fidelity events. 
In summary, rendering temporally dense, high-resolution event streams that match the $\mu$s precision of event cameras in real time remains challenging.

\vspace{-3pt}
\subsection{Spiking Neural Network (SNN) for Event-based Vision}
The concept of SNN has been explored for over two decades~\cite{gerstner1995time, maass1997networks}. As a bio-inspired architecture that mimics neuron activities in human brains, SNNs are well-suited for processing spatiotemporal data~\cite{tavanaei2019deep} at a reduced computation cost than ordinary artificial neural networks (ANNs)~\cite{merolla2014million, pei2019towards}. 
Since spiking units such as variants of leaky integrate-and-fire (LIF)~\cite{gerstner2014neuronal} are non-differentiable, various strategies to back propagate through SNNs have been developed, including the prevalent ANN-to-SNN conversion~\cite{hunsberger2015spiking, deng2021optimal} and the surrogate gradient method~\cite{wu2018spatio, neftci2019surrogate}.
As a result, SNN is becoming increasing popular in computer vision and deep learning in general~\cite{schuman2022opportunities, rathi2023exploring}.

The neuromorphic design and low-cost properties of SNNs  naturally draw researchers towards applying them on event streams. 
A hybrid ANN-SNN architecture is usually employed to process event data into desired outputs~\cite{kugele2021hybrid, ahmed2024hybrid}, and SNN-empowered methods have demonstrated success across multiple downstream tasks, including video reconstruction~\cite{zhu2022event}, classification~\cite{yao2021temporal}, action recognition~\cite{liu2021event}, and optical flow estimation~\cite{hagenaars2021self}. To the best of our knowledge, \method pioneers in leveraging spiking units to \emph{generate}, instead of process, event streams.

\section{Method} \label{sec:method}
\subsection{Motivation: Noise Factors in Event Sensory}
Unlike conventional frame-based sensors that record absolute pixel illuminance via exposure, event sensors constantly monitor each pixel's illuminance and fire a spike (\textit{i.e.,} an event) whenever its logarithmic change exceeds a predefined contrast threshold $\rho$. 
As a result, an event camera outputs a sequence of asynchronous spikes, each denoted by a tuple $(t, x, y, p)$, where $t$ denotes the timestamp, $(x, y)$ are the pixel coordinates, and $p \in \{+1, -1\}$ indicates the polarity of the change (positive or negative). 
This design grants event sensors advantages such as extremely high temporal resolution ($<$100 $\mu$s) and high dynamic range, but it also introduces unique noise compositions to the resulting event streams. Realistically reproducing the distribution of real-world events has thus become a key challenge of event synthesis.

\begin{figure}
    \centering
    \includegraphics[width=0.65\linewidth]{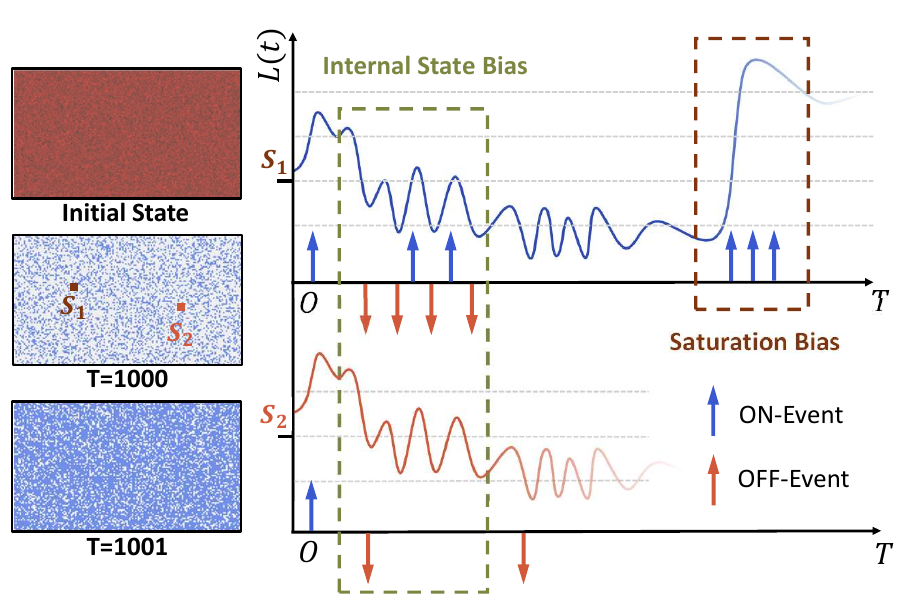}
    \vspace{-0.3cm}
    \caption{\emph{Left}: (top) Each pixel is initialized with a randomly sampled internal state; (bottom) Identical brightness changes are applied to all pixels, causing them to fire events at different timestamps. 
    \emph{Right}: {\color{isb}\textbf{Internal State Bias}}: Two event pixels with different initial states ($S_1$ and $S_2$) produce drastically different event pulses ({\color{blueevent}\textbf{blue}} and {\color{redevent}\textbf{red}} arrows) even when they undergo identical illuminance changes; {\color{sb}\textbf{Saturation Bias}}: instant changes in illuminance create trailing events afterwards.}
    \vspace{-9pt}
    \label{fig:threshold_bias}
\end{figure}

\paragraph{Internal state bias}
The asynchronous nature of event sensor causes its pixel circuits to have different \emph{voltage levels
(\textit{ i.e.,} internal states)} at each moment. In other words, two event pixels may produce drastically different events during a given time interval, even when their illuminance changes are identical, as illustrated in Fig.~\ref{fig:threshold_bias}. Such an internal state bias is the main factor why real-world event streams appear noisy to human eyes.

As a result, inferring event occurrences solely from the illuminance difference between two consecutive frames, a common practice undertaken by existing event simulators~\cite{rebecqESIMOpenEvent2018}, almost certainly results in suboptimal performance. To resolve this issue, our proposed \model network scans through the entire temporal span of the input clip and keeps the internal state recorded within its BiLIF activation unit. This approach realistically simulates the behavior of event cameras and ensures a consistent distribution of events across the entire video.

\paragraph{Saturation effect}
After firing an event spike, the event pixel undergoes a reset period before it can give further responses. Therefore, when the instant illumination change is very large, it will take the sensor a relatively long time to fire all events, which creates a sequence of pulses in the observed event stream, as visualized in Fig.~\ref{fig:threshold_bias}. We call this phenomenon the \emph{saturation effect}. Failing to address such an effect in simulation will increase the domain gap between synthetic and real-world events, especially around extremely bright or extremely dark regions, and further impair the utility of generated event data on downstream tasks.
While this effect has been addressed by some physics-based simulation methods~\cite{huV2eVideoFrames2021}, reproducing it with artificial neural network (ANN) modules would require a large receptive field and longer training time, and thus it is often ignored by existing learning-based approaches~\cite{zhangV2CEVideoContinuous2024}. To this end, we develop the \emph{bipolar leaky integrate-and-fire (BiLIF) spiking unit} that naturally exhibits the saturation effect without explicit training.


\paragraph{Other noise factors}
Beyond threshold bias and saturation, event streams also experience minor perturbations from event camera circuitry, such as threshold mismatch, hot pixels, leak noise events, and shot noise \cite{huV2eVideoFrames2021}. In our pipeline, low-SPP Monte Carlo path tracing naturally introduces shot-noise–like variations in the RGB input, and our \model network also helps to emulate these effects by learning from the outputs of the physics-based V2E method, resulting in synthesized events that closely match the noise characteristics of real sensors.

\begin{figure*}[t]
    \centering
    \includegraphics[width=\linewidth]{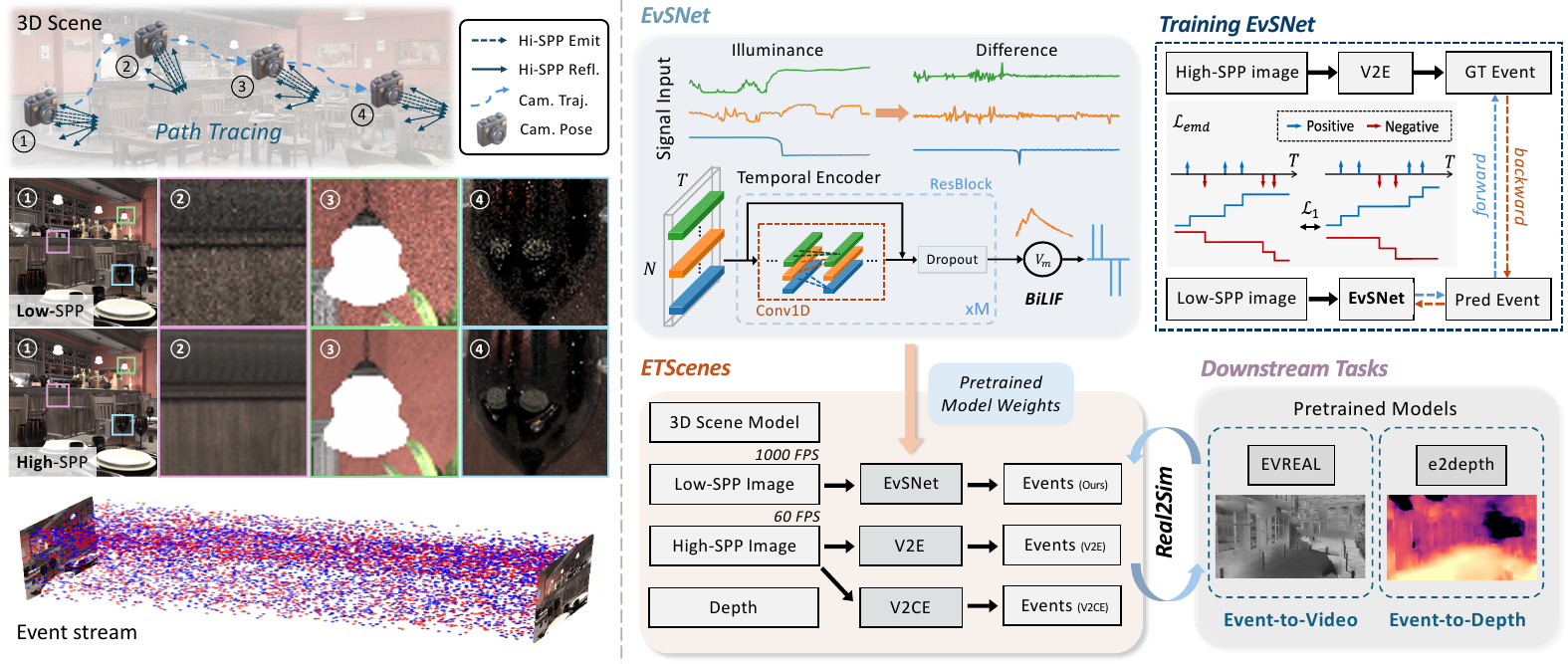}
    \vspace{-18pt}
    \caption{Pipeline of our \emph{\method}.
     \textit{Bottom left}: Our task is to directly render event streams from 3D scenes. \textit{Top-left}: In each scene, we define a camera trajectory and run the path tracer at varying sample-per-pixel (SPP) to obtain low-SPP (noisy, high FPS) and high-SPP (noiseless, low FPS) RGB videos. 
    \textit{Top-right}: Our \model network takes noisy pixel illuminances as input and outputs discrete event spikes via the BiLIF unit; it is trained with V2E-simulated events, using an earth mover's distance (EMD) loss. 
    \textit{Bottom-right}: While existing event simulators (e.g., V2E, V2CE) can only run on high-SPP videos, \model is able to convert low-SPP inputs to event streams. We validate its functionality through the \emph{Real2Sim} test, conducted on two downstream tasks.
    }
    \label{fig:pipeline}
    \vspace{-3pt}
\end{figure*}

\subsection{\method: Path Tracing-based Event Rendering}
To synthesize event streams from 3D scenes, we firstly discretize time over a duration of $T$ seconds with a high frame rate $F_{\mathrm{evt}}$ into $K = T \cdot F_{\mathrm{evt}}$ timestamps. Then for each $t_k = k / F_{\mathrm{evt}}$, we render an RGB image $\mathbf{I}_k$ with \emph{Monte Carlo path tracing}. While various other global illumination (GI) methods—such as Lumen in UE5, DLSS and NRD—may be used for real-time, noise-free rendering, their temporal accumulation strategies tend to blur or suppress edge features in scenes with fast-moving objects, making them suboptimal choices for simulating event streams that are particularly sensitive to edge movements.

To make sure simulated event data has sufficiently high temporal resolution, we choose $F_{\mathrm{evt}} = 1,000$. 
However, conventional high sample-per-pixel (SPP) path tracing (at least $N = 2,048$ SPP is needed to produce noiseless RGB frames) is impractical under this setting due to its excessive time consumption. Therefore, we decrease the SPP to $N = 64$ in our pipeline, reducing the runtime of the path tracer to $1/32$. The resulting sequence $\{\mathbf{I}_0, \mathbf{I}_1, \dots, \mathbf{I}_{K}\}$ is then a high-FPS yet noisy approximation of real-world capture. As simply computing events by frame difference cannot account for the various noise factors in real-world event streams, we instead train a \emph{pixel-wise event pulsing network (\model)} to simultaneously denoise path tracing results and predict event spikes. Specifically, for each pixel location $(x, y)$, we extract the sequence of pixel values $\{I_0^{(x,y)}, \dots, I_K^{(x,y)}\}$ and convert it to logarithmic differences of illuminance as follows:
\vspace{-2pt}
\begin{align}
\begin{split}
    L &= 0.2126 \, I_R + 0.7152 \, I_G + 0.0722 \, I_B \, , \\
    L^{\mathrm{log}} &= \begin{cases}
        \ln L \, , \, & L \ge \rho\\
        \frac{\ln\rho}{\rho}L \, ,\, & L < \rho
      \end{cases}, \quad X_k = L^{\mathrm{log}}_{k} - L^{\mathrm{log}}_{k - 1}.
\end{split}
\end{align}
Here $\rho$ is a predefined threshold and we omit the pixel location $(x, y)$ for clarity. We then input the processed sequence $\mathbf{X} = \{X_1, \dots, X_{K}\}$ into \model and obtain simulated events $\mathbf{E} = \{e_1, \dots, e_{K }\}$. More details on \model will be explained in Sec.~\ref{sec:network}.


Since \model performs pixel-wise and temporally local (with receptive field width $W < 50$) inference, we are able to integrate it into the Falcor~\cite{Kallweit22} renderer using the NVIDIA TensorRT framework~\cite{tensorrt} and easily parallelize the computation across the whole frame, achieving a rendering speed of $\sim$ 4 minutes per second of 720p events video. The overall pipeline of \method is visualized in Fig.~\ref{fig:pipeline}.

\subsection{\model: Lightweight Event Spiking Network}
\label{sec:network}

\paragraph{Spiking neuron preliminaries.}
Spiking neurons are special activation units that convert sequential inputs into discrete spikes when the membrane potential crosses a certain threshold.
A commonly used spiking neuron is the leaky integrate-and-fire (LIF) model~\cite{gerstner2014neuronal}, whose dynamics are described as:
\begin{align} \label{eq:snn-origin}
\begin{split}
    \textbf{(charge) }\; &V'(t) = \, (1 - 1/\tau)\cdot V(t - 1) + I(t),\\
    \textbf{(fire) }\; &S(t) = \mathbb{1}\bigl[V'(t) \ge V_{\mathrm{th}}\bigr],\\
    \textbf{(reset) }\; &V(t) = V'(t) - S(t),
\end{split}
\end{align}
where $V(t)$ is the membrane potential, $I(t)$ is the synaptic input current, $S(t)$ is the output spike, and $\tau$ is the decay constant. In a word, an output spike is generated when $V(t)$ exceeds the threshold $V_{\mathrm{th}}$, after which $V(t)$ is reset. To enable gradient‐based training, a surrogate derivative can be used for the non‐differentiable spike function~\cite{neftci2019surrogate}.

\paragraph{Bipolar LIF (BiLIF) for event simulation}
To accommodate the inherently bipolar nature of event camera outputs, we extend the conventional LIF neuron to a \emph{bipolar LIF}, dubbed BiLIF, that is able to emit both positive and negative event spikes. In BiLIF, the charging and resetting of membrane potential $V(t)$ follows the formulation in \cref{eq:snn-origin}, while the firing step is now governed by two symmetric thresholds, $+V_{\mathrm{th}}$ and $-V_{\mathrm{th}}$. Specifically, the bipolar spike $S(t)\in\{-1,0,+1\}$ is computed according to:
\begin{equation}
  S(t) = \operatorname{sgn}\bigl[V(t)\bigr]\,\mathbb{1}\bigl(|V(t)| \ge V_{\mathrm{th}}\bigr)\, ,
\end{equation}
where $\operatorname{sgn}(u)=+1$ if $u>0$ and $-1$ if $u<0\,$.

This mechanism allows the neuron to simultaneously represent increases and decreases in illuminance as positive and negative pulses, respectively. 
Similar to the vanilla LIF, we also adopt a surrogate gradient approach for the non‐differentiable Heaviside function. Let $\sigma(u)$ be a smooth surrogate function, during backpropagation we can approximate that:
\vspace{-3pt}
\begin{align}
    \frac{\partial S(t)}{\partial V(t)} \approx \frac{\partial\sigma\bigl(V(t)-V_{\mathrm{th}}\bigr)}{\partial V(t)} - \frac{\partial\sigma\bigl(-V(t)-V_{\mathrm{th}}\bigr)}{\partial V(t)}\,.
\end{align}
Our training uses the fast arctangent surrogate~\cite{aliyev2024fine}. This formulation ensures both ON and OFF events contribute meaningful gradients, enabling the network to learn representations that capture the full polarity spectrum of event streams.
\paragraph{Network architecture}
We design our \model network to be lightweight, pixel-wise, and temporally local so that it can be smoothly integrated into the rendering pipeline. 
As previously mentioned, it takes as input a sequence $\mathbf{X}$ of logarithmic differences of illuminance signals. 
Then it passes $\mathbf{X}$ through an 1D convolutional encoder to extract per-timestamp features:
\begin{align}
\begin{split}
    \mathbf{h}^{(0)} &= \mathrm{ReLU}\bigl[\mathrm{Conv1d}(\mathbf{X})\bigr],\\
    \mathbf{h}^{(i)} &= \mathrm{ReLU}\Bigl[\mathbf{h}^{(i-1)} + \mathrm{Conv1d}\Bigl(\mathrm{ReLU}\bigl[\mathrm{Conv1d}(\mathbf{h^{(i-1)}})\bigr]\Bigr)\Bigr],
\end{split}
\end{align}
where $i = 1, \dots, M$. Finally, a point-wise convolution is used to project the feature tensor to a length-$K$ sequence that is subsequently used to excite event pulses via a BiLIF spiking unit:
\begin{equation}
\mathbf{S} = \mathrm{BiLIF}\bigl[\mathrm{Conv1d}\bigl(\mathbf{h}^{(M)}\bigr)\bigr]\in\{-1,0,1\}^{B\times K}\,.
\end{equation}

The temporal receptive field of \model is relatively narrow, and thus the renderer only needs to refer to a small set of neighboring frames to infer the event distribution at the current timestamp. This design effectively reduces the memory and time consumption of the \method pipeline.



\paragraph{Loss functions and training.}
We render high-quality, $F_{\mathrm{evt}}$ FPS videos with high-SPP path tracing and feed them into the V2E toolbox~\cite{huV2eVideoFrames2021} to generate ground truth events $\mathbf{E}$. To align the predicted event sequence $\mathbf{S}$ with $\mathbf{E}$, both of which are sparse and discrete-valued, we construct the loss as the 1D earth mover's distance (EMD) between them. It is straightforward to deduce that the EMD between two equal-length 1D sequences is the pairwise distance between their sorted entries~\cite{villani2008optimal}. Therefore, the loss can be computed as:
\vspace{-2pt}
\begin{align}
    \mathcal{L}(\mathbf{E}, \mathbf{S}) 
    = \frac{1}{K}\sum_{i=1}^K\Bigl\lVert\sum_{j=1}^i\mathbf{S}_j - \sum_{j=1}^i\mathbf{E}_j\Bigr\rVert_{1},
    \label{eq:emd}
\end{align}
that is, the difference between the cumulative sums of $\mathbf{S}$ and $\mathbf{E}$ at each timestamp in the sequence.
However, this formulation presumes that $\mathbf{E}$ and $\mathbf{S}$ have the same number of nonzero entries. While this condition is not inherently met by the \model network, Eq.~\ref{eq:emd} will prompt the model to hallucinate nonexistent events towards the end of the sequence. To address this drawback, we compute the EMD twice in different directions as:
\vspace{-3pt}
\begin{equation}
    \mathcal{L}'(\mathbf{E}, \mathbf{S})  = \frac{1}{2}\bigl[\mathcal{L}(\mathbf{E}, \mathbf{S}) + \mathcal{L}(\mathbf{E}^{\mathrm{rev}}, \mathbf{S}^{\mathrm{rev}})\bigr] \, ,
    \label{eq:emd2}
\end{equation}
where $\mathbf{E}^{\mathrm{rev}}$ and $\mathbf{S}^{\mathrm{rev}}$ are the reversed sequences.

Considering that $\mathbf{E}$ and $\mathbf{S}$ contain both positive and negative values, we separate them into two channels and apply ~\cref{eq:emd2} individually to each channel:
\vspace{-3pt}
\begin{equation}
    \mathcal{L}_{\mathrm{EMD}}(\mathbf{E}, \mathbf{S}) = \mathcal{L}'\Big(\mathbf{E}_{>0}, \mathbf{S}_{>0}\cdot \mathbf{S}\Big) + \mathcal{L}'\Big(\mathbf{E}_{<0}, \mathbf{S}_{<0}\cdot(-\mathbf{S})\Big).
\end{equation}

To further constrain \model on the total number of event spikes, we additionally regularize training with a event spike count loss:
\vspace{-3pt}
\begin{equation}
\mathcal{L}_{\mathrm{count}} = \Big\|\sum_{i=1}^K |\mathbf{S}_i| - \sum_{i=1}^K|\mathbf{E}_i|\Big\|_1.
\end{equation}

The total loss is then the sum of two terms with a weight $\lambda$:
\vspace{-3pt}
\begin{equation}
    \mathcal{L}_{\mathrm{total}} = \mathcal{L}_{\mathrm{EMD}} + \lambda\cdot\mathcal{L}_{\mathrm{count}}.
\end{equation}

\begin{figure*}
    \centering
    \includegraphics[width=\linewidth]{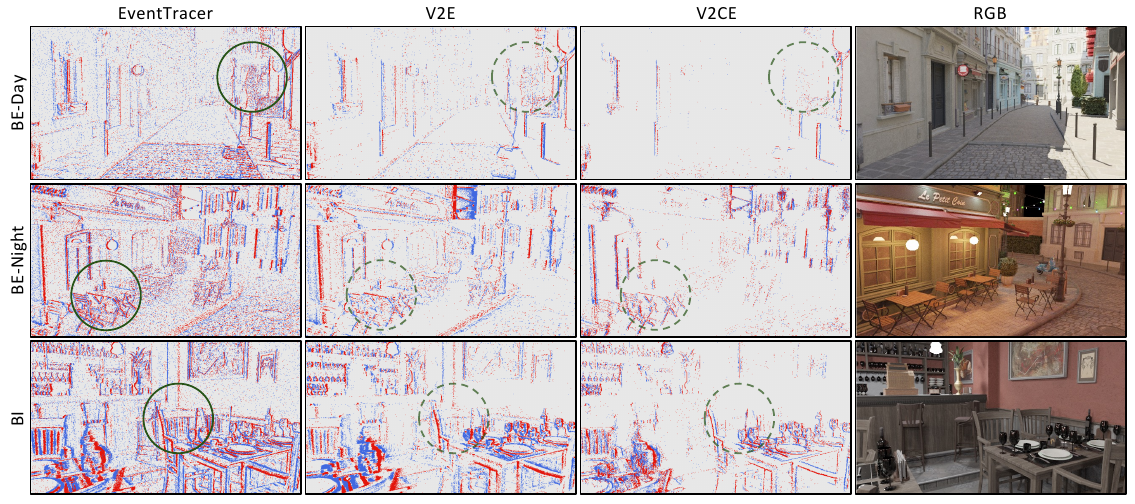}
    \caption{Comparison of events obtained using various rendering approaches--\method, V2E, and V2CE--on three scenes: Bistro-Exterior-Day (BE-Day), Bistro-Exterior-Night (BE-Night) and Bistro-Interior (BI). We also provide the corresponding RGB frames for reference.}
    \label{fig:events-comparison}
\end{figure*}

\section{Assessment} \label{sec:results}

\subsection{Implementation Details}

We curate the training data of \emph{\model} by rendering low- and high-SPP videos paired with V2E-generated events from Amazon Lumberyard Bistro~\cite{ORCAAmazonBistro}. For each of the three scenes, Bistro-Interior, Bistro-Exterior-Day and Bistro-Exterior-Night, we render 2$\sim$4 seconds of video with $F_{\mathrm{evt}} = 1,000$ FPS, amounting to 7,936 frames in total. While constructing the training set alone takes $>$12 hours, the trained \model network and the overall \method pipeline can be readily applied to unseen data, reducing its amortized time consumption for each new scene rendered.

We set the kernel size for all but the last convolution layers to 7 and the network depth $M = 3$, resulting in a relatively small temporal receptive field width of $W = 43$. BiLIF is implemented using the SpikingJelly framework~\cite{spikingjelly}. The model is trained for 20 epochs and all experiments are conducted on a NVIDIA GeForce RTX 4090 GPU. The resolution of rendered images is $1,280\times720$, while for downstream tasks we downsample it to $640\times360$, since most existing baselines and pretrained models are tuned for low-resolution inputs. \textbf{More implementation details can be found in the Supplementary Material.}

\paragraph{The \dataset dataset}
After training the \model network, we construct the \emph{\dataset} dataset with paired event-RGB videos to validate our \method pipeline. Given a dynamic 3D scene, we render event stream with \method and its corresponding RGB video with regular high SPP path tracing at a frame rate of 60 FPS. A total of 6 scenes are used: three from Amazon Lumberyard Bistro and three from \cite{resources16}, namely Kitchen, Staircase and Classroom. We render 20 seconds of videos for Amazon Lumberyard Bistro scenes, and 10 seconds for the remaining scenes, resulting in an overall 90 seconds of videos with $\sim$6,000 RGB frames and (equivalently) $\sim$90,000 event ``frames''. Additionally, we also render ground truth depth maps corresponding to each RGB frame with almost no extra computation cost. We note that other annotations, such as surface normals, optical flows and segmentation masks, can be easily obtained likewise.

\subsection{Validation Setup}
Quantitatively measuring event stream's fidelity can be challenging, as raw events are not readily interpretable or visually understandable to humans~\cite{ercan2023evreal}. 
To better assess \dataset fidelity and compare \method against other event simulators, we select two downstream tasks, \emph{event-to-video reconstruction (E2V)} and \emph{event-to-depth estimation (E2D)}, and conduct the \emph{Real2Sim} test for each task. Specifically, we evaluate baseline models pretrained with existing real-world data $\mathcal{D}_{\mathrm{gt}}$ on different simulated datasets $\mathcal{D}_{\mathrm{sim}}$. Better results indicate a greater similarity between $\mathcal{D}_{\mathrm{gt}}$ and $\mathcal{D}_{\mathrm{sim}}$.

\paragraph{Event-to-video reconstruction (E2V)}
One of the most fundamental tasks of event-based vision, event-to-video aims at reconstructing frame‐based intensity images from input event streams. We adopt the EVREAL benchmark~\cite{ercan2023evreal} for the \emph{Real2Sim} test, which supports the evaluation of 8 baseline E2V models on the given data using multiple quantitative metrics.

\paragraph{Event-to-depth reconstruction (E2D)}
Recovering per‐pixel depth maps directly from asynchronous event streams has emerged as a key challenge in neuromorphic vision, promising robust range sensing under high‐speed and high dynamic range conditions. 
We perform \emph{Real2Sim} validation using the e2depth model~\cite{hidalgo2020learning} pretrained on mixed DENSE (synthetic) and MVSEC (real)~\cite{zhu2018multivehicle} datasets.


\paragraph{Comparison to other event simulators}
We compare \method against two mainstream event simulators, V2E~\cite{huV2eVideoFrames2021} and V2CE~\cite{zhangV2CEVideoContinuous2024}, and render the counterparts of \dataset using each method, denoted as \dataset-V2E and \dataset-V2CE, respectively, from RGB videos whose frame rate does not exceed 60 FPS for a fair comparison. Specifically, V2CE has a fixed internal threshold and cannot simulate events when adjacent frames are too similar; therefore, we downsample the input videos to 15 FPS so that \dataset-V2CE contains meaningful event streams. A qualitative comparison across three renderings is available in Fig.~\ref{fig:events-comparison}, where we can readily observe that \method preserves more scene details, especially high-frequency patterns that produce quickly alternating event signals as the camera moves.

\begin{figure}[htbp]
    \centering
    \includegraphics[width=0.65\linewidth]{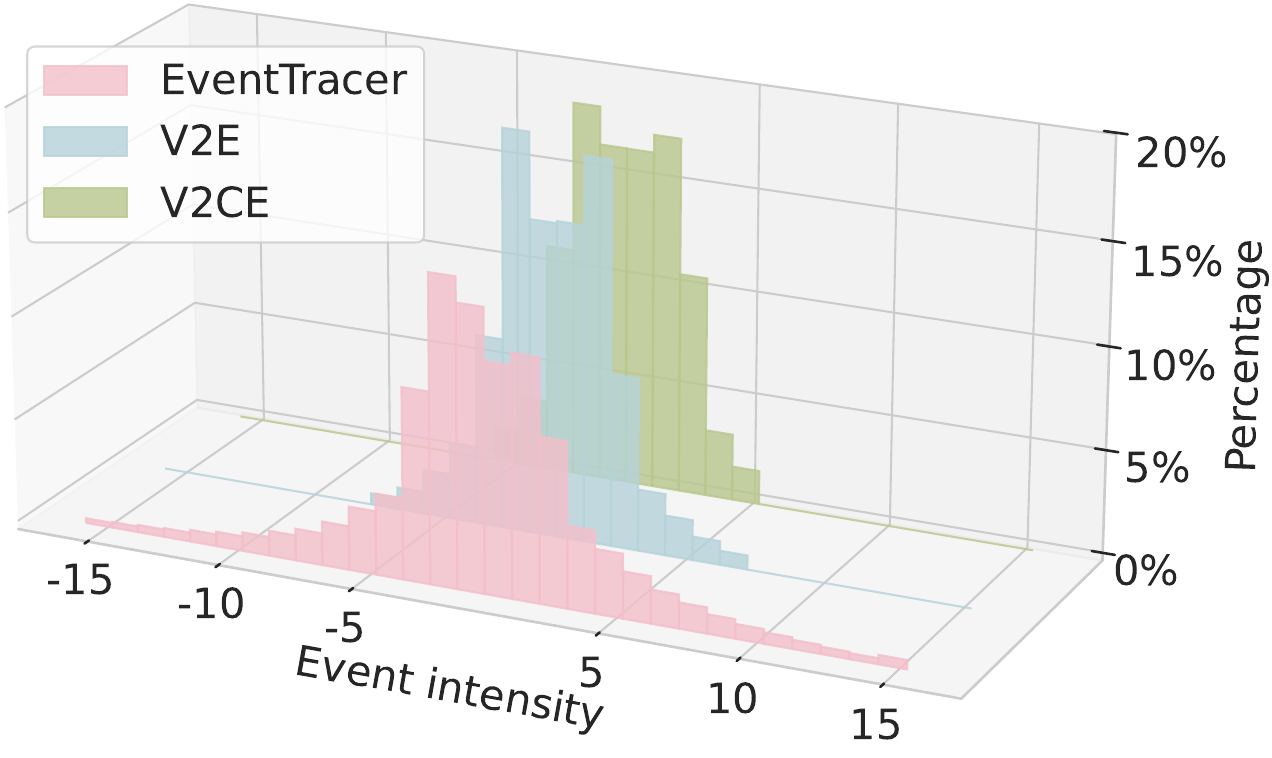}
    \caption{Histograms of event intensity generated by three investigated simulators. We take all frames in the BistroExterior-Day scene and integrate event streams to 60 FPS voxels. }
    \label{fig:evt_dist}
\end{figure}

We also visualize the histogram of event intensity in each dataset in Fig.~\ref{fig:evt_dist}. We find that \dataset features a more balanced distribution, with a good number of pixels having high event intensity ($\geq$10 events), faithfully reflecting the movements of sharp edges and the presence of bright light sources in the scene. In contrast, event frames integrated from \dataset-V2E and \dataset-V2CE contain mostly weakly activated pixels, making the modeling of high-contrast scene contents exceptionally difficult.

\subsection{Event-to-Video Reconstruction}
\begin{table*}[htbp]
\centering
\footnotesize
\caption{Quantitative comparison under the \emph{Real2Sim} setting using multiple E2V baselines. Metrics (from left to right): SSIM$\uparrow$, LPIPS$\downarrow$ and LPA ($\times 10^3$)$\uparrow$.}
\label{tab:pre_e2v}
\adjustbox{width=\linewidth}{
\begin{tabular}{l|c|c|c|c|c|c|c|c}
\toprule
Methods & E2VID & E2VID+ & FireNet & FireNet+ & SPADE–E2VID & SSL–E2VID & ET–Net & HyperE2VID \\
\midrule
V2CE & 0.244 / 0.648 / 0.23 & 0.238 / 0.513 / \textbf{0.44} & 0.215 / 0.607 / 0.19 & 0.218 / 0.578 / 0.74 & 0.242 / 0.550 / 0.16 & 0.193 / 0.705 / 0.14 & 0.238 / 0.525 / \textbf{0.50} & 0.240 / 0.518 / 0.66 \\
V2E & \textbf{0.351} / \textbf{0.501} / 0.14 & 0.395 / 0.410 / 0.18 & 0.370 / 0.472 / 0.09 & 0.326 / 0.457 / 0.58 & 0.356 / 0.421 / 0.15 & 0.202 / 0.674 / 0.06 & 0.398 / 0.400 / 0.25 & 0.393 / 0.395 / 0.33 \\
\midrule 
\method & 0.348 / 0.478/ \textbf{0.24} & \textbf{0.413} / \textbf{0.375} / 0.32 & \textbf{0.424} / \textbf{0.400} / \textbf{0.42} & \textbf{0.384} / \textbf{0.399} / \textbf{1.16} & \textbf{0.405} / \textbf{0.404} / \textbf{0.40} & \textbf{0.289} / \textbf{0.605} / \textbf{0.29} & \textbf{0.409} / \textbf{0.387} / {0.42} & \textbf{0.438} / \textbf{0.350} / \textbf{0.70} \\
\bottomrule
\end{tabular}
}
\end{table*}

\begin{figure*}[htbp]
    \centering
    \includegraphics[width=\linewidth]{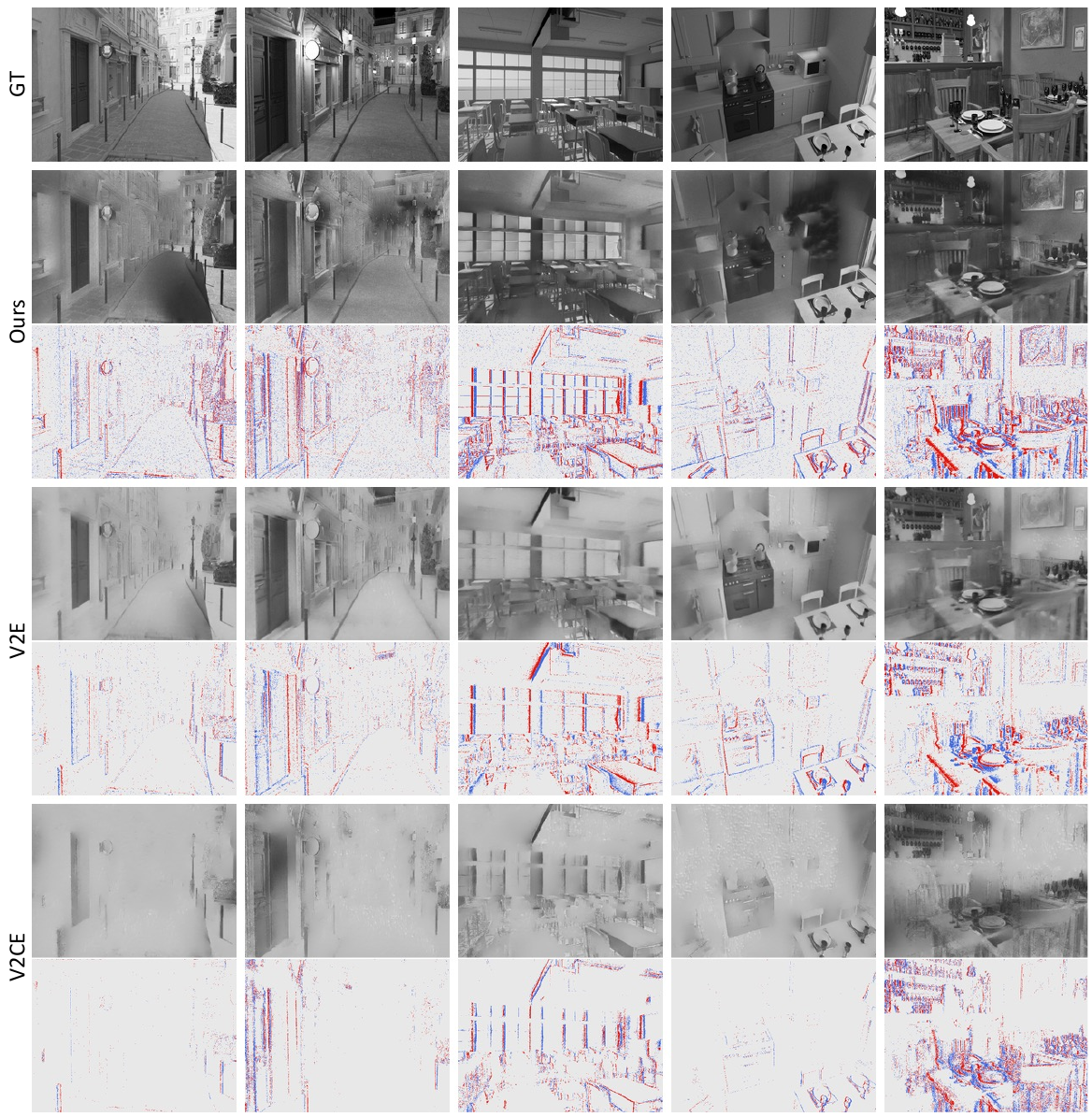}
    \caption{\emph{Real2Sim} validation results on E2V. Each column contains the ground truth grayscale frame as well as pairs of reconstructed image and event frame for each event simulator. When evaluated on our \dataset dataset, the pretrained FireNet~\cite{scheerlinck2020fast} model is able to reconstruct more details in the scene, including lamp posts, window grids, chairs, and paintings. In contrast, evaluation results on \dataset-V2E and \dataset-V2CE contain large blank regions, suggesting information loss in their event data.}
    \label{fig:E2V_result_1}
\end{figure*}

\begin{figure*}[htbp]
    \centering
    \includegraphics[width=\linewidth]{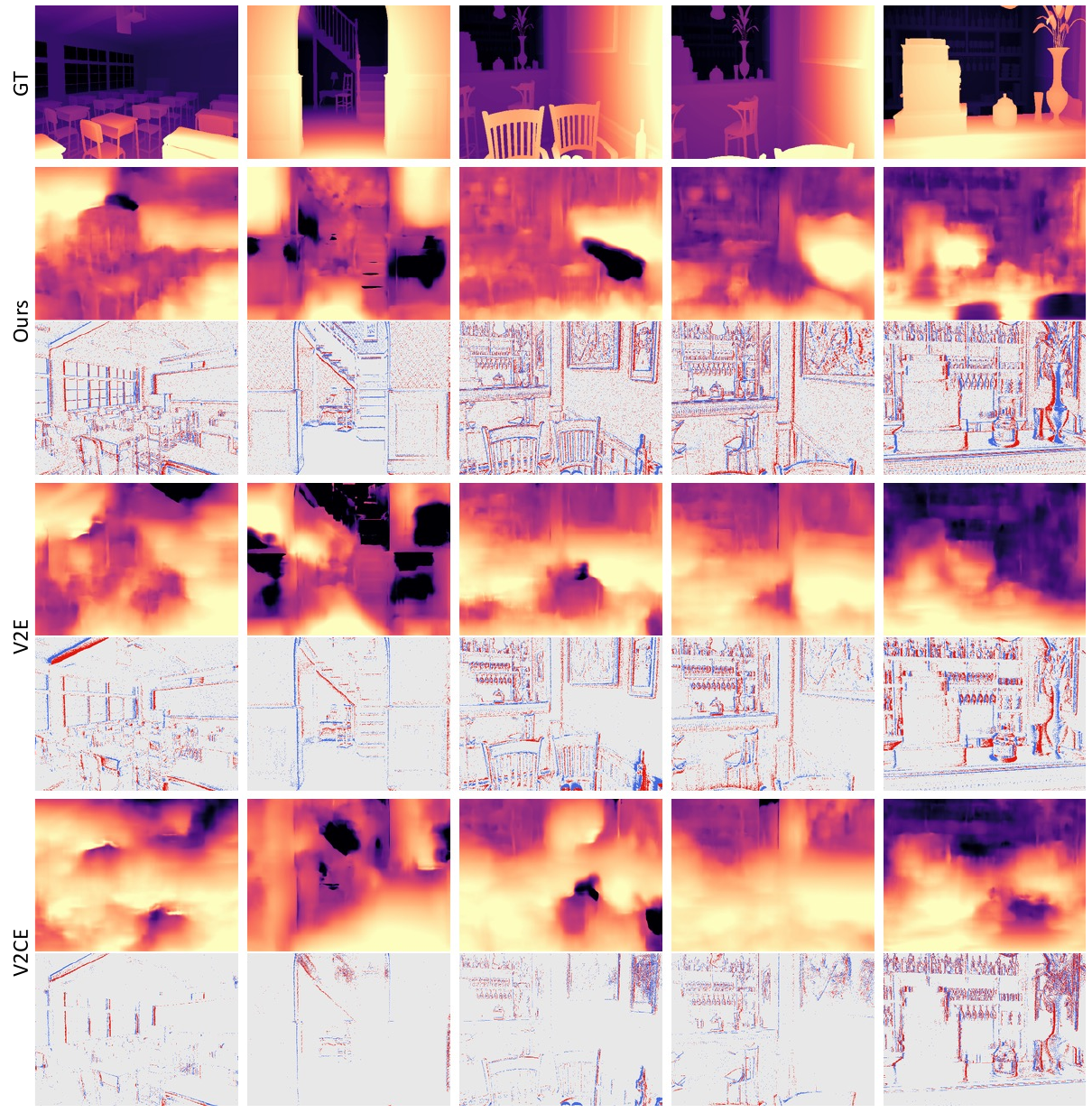}
    \caption{\emph{Real2Sim} validation results on E2D. Each column contains the ground truth depth map as well as pairs of estimated depth and event frame for each event simulator. 
    Note that \dataset yields the best results of three, where not only detailed structures (\textit{e.g.} desks, stairs, and bar counters) are visible, but the overall depth estimation for each scene also displays certain consistency.
    When using \dataset-V2E and \dataset-V2CE, more details are lost and scene layouts are barely distinguishable from the output depth maps.
    }
    \label{fig:E2D_result_1}
\end{figure*}

Since event streams only record relative changes in illuminance, the illuminance level of videos reconstructed by E2V models can vary from case to case, largely affecting the scores of various metrics and making them less indicative of the actual image quality. To resolve this issue, we apply histogram equalization (HE) on both ground truth and reconstructed RGB frames to map their overall illuminance to the same level. 
We report structural similarity (SSIM), learned perceptual image patch similarity (LPIPS)~\cite{zhang2018unreasonable}, and energy of image Laplacian (LPA)~\cite{subbarao1993focusing} between reconstructed images and ground truth after HE. We refrain from using pixel-wise intensity metrics such as mean square error (MSE) or peak signal-to-noise ratio (PSNR) due to their high sensitivity to image illuminance level.

Table~\ref{tab:pre_e2v} compares \dataset, \dataset-V2E, and \dataset-V2CE quantitatively in \emph{Real2Sim}. It can be observed that events simulated by \method align better with models pretrained on real-world data, resulting in higher-quality and sharper video reconstruction. A quantitative visualization is provided in Fig.~\ref{fig:E2V_result_1}, where \method demonstrates its ability to retain high-frequency features such as street tiles and window grids, while the reconstructed images from \dataset-V2E and \dataset-V2CE exhibit large featureless regions due to the loss in visual details.



\subsection{Event-to-Depth Estimation}

The depth distribution of our \dataset dataset, featuring mostly indoor scenes, is drastically different from that of autonomous driving datasets such as DENSE and MVSEC. Therefore, we only evaluate relative error metrics to quantify the performance of e2depth models, including absolute relative error (Abs.Rel.), square relative error (Sq.Rel.), linear RMSE (RMSE), logarithmic RMSE (RMSE log), and scale‐invariant logarithmic error (SI.log).

Table~\ref{tab:pre_e2d} compares the \emph{Real2Sim} performance of different event simulators. Our \method achieves better scores in all three metrics, again demonstrating its capability to produce realistic event streams that can be readily processed by pretrained deep models. A qualitative comparison is available in Fig.~\ref{fig:E2D_result_1}. We can observe that the pretrained e2depth model identifies more scene layouts, such as desks, stairs and bar counters, and makes more consistent depth estimations for those objects when evaluated on \dataset.



\subsection{Rendering Speed}

Our \model runs at a speed of 22.85ms per inference, and it takes \method 1.12 minutes to generate one second of 360p event video (mpsv). In comparison, high-SPP RGB rendering takes 1.46 mpsv and 0.37 mpsv for V2E (60 FPS) and V2CE (15 FPS), and their event simulation speeds (including runtime for frame interpolation models) are 0.39 mpsv and 0.05 mpsv, which sum up to 1.85 mpsv and 0.42 mpsv of overall generating speed, respectively. Were 60 FPS inputs are used for V2CE, its speed would be degraded accordingly to $>$1.5 mpsv, more than 30\% slower than \method.

\begin{table}[htbp]
\centering
\footnotesize
\caption{Quantitative comparison under the \emph{Real2Sim} setting using pretrained \emph{e2depth} network.}
\label{tab:pre_e2d}
\adjustbox{width=0.7\linewidth}{
\begin{tabular}{l|c|c|c|c|c}
\toprule
Method & Abs. Rel. $\downarrow$ & Sq. Rel. $\downarrow$ & RMSE $\downarrow$ & RMSE log $\downarrow$ & SI. log $\downarrow$ \\
\midrule
V2CE & 0.898 & 6.304 & 29.172 & 1.393 & 0.582 \\
V2E & 0.990 & 9.917 & 28.225 & 1.284 & 0.529 \\
\midrule
\method & \textbf{0.896} & \textbf{5.550} & 
\textbf{28.17} & 
\textbf{1.147} & 
\textbf{0.485} \\
\bottomrule
\end{tabular}
}
\end{table}

\section{Conclusion} 
\label{sec:conclusion}

In this work, we have presented a novel path tracing-based event rendering pipeline, dubbed \method, realizing fast and high-fidelity simulation of event data from 3D scenes.
This pipeline consists of an efficient low-SPP Monte Carlo path tracing renderer and a lightweight event spiking network, featuring the BiLIF spiking unit and trained with the earth mover's distance loss to capture the physical properties of event signals.
We compare \method against existing event simulators by constructing synthetic event-RGB datasets and evaluating them on two downstream tasks with the \emph{Real2Sim} test. Experiment results demonstrate that \method outperforms existing tools such as V2E and V2CE in both event stream fidelity and rendering speed, thereby narrowing the sim-to-real gap that hinders a wide range of event-based vision research.

\paragraph{Limitations and future works.}
While our \model network can be trained to approximate the pixel-wise event distribution with high fidelity, it is yet only weakly constrained to produce the same number of event pulses as the ground-truth signal. As a result, its outputs may induce degraded performance in downstream tasks that are sensitive to event counts, for example, event-to-video reconstruction. It is left for future works to explore additional loss terms and/or training strategies to mitigate this issue.
Besides, as a fully differentiable and path tracing-based solution to event synthesis, \method also inspires us to explore the potential of fusing event data with \emph{differentiable optics}, where researchers develop designated point spread functions (PSFs) and subsequently utilize these PSFs to encode target images~\cite{yang2024curriculum,shi2024learned}. Advances in this direction will equip event sensors with better imaging quality and improved sensitivity to depth and motion, ultimately reshaping the field of event-based vision.

\paragraph{Bells \& whistles: Sim2Real}
In addition to \emph{Real2Sim}, we may also validate the performance of event simulators via \emph{Sim2Real}, namely to train baseline models with simulated event-RGB data and to evaluate on real-world datasets. However, the scale and diversity of our \dataset dataset are still insufficient for training a fully functional V2E or V2D network. The domain gap between simple, mostly static scenes in \dataset and real-world data with sophisticated lighting conditions and object motion also leads to low-quality results on datasets such as UZH DAVIS~\cite{mueggler2017event} and MVSEC~\cite{zhuMultiVehicleStereo2018}. \textbf{Please refer to the Supplementary Material for more detailsand preliminary results.} Yet we believe that such a gap can be narrowed by applying \method on scenes with more complex layouts and under specialized settings such as autonomous driving, creating a larger-scale simulated event-RGB dataset suitable for a broader range of downstream tasks. We consider this an important step for follow-up research.

\bibliographystyle{unsrt}
\bibliography{reference}

\end{document}